\documentclass{article} % For LaTeX2e
\usepackage{iclr2024_conference,times}

\usepackage{amsmath,amsfonts,bm}

\def\Figref#1{Figure~\ref{#1}}

\def\secref#1{section~\ref{#1}}
\def\eqref#1{equation~\ref{#1}}
\def\1{\bm{1}}

\DeclareMathAlphabet{\mathsfit}{\encodingdefault}{\sfdefault}{m}{sl}
\SetMathAlphabet{\mathsfit}{bold}{\encodingdefault}{\sfdefault}{bx}{n}

\usepackage{hyperref}
\usepackage{url}
\usepackage{tikz}
\usetikzlibrary{decorations.pathreplacing,shapes.geometric}
\usepackage{booktabs}
\usepackage{multirow}
\usepackage[most]{tcolorbox}

\usepackage{xcolor}

\usepackage{siunitx}

\usepackage{makecell}
\usepackage{lscape}
\usepackage{subcaption}

\title{LAB: Large-Scale Alignment for ChatBots }

\author{MIT-IBM Watson AI Lab and IBM Research \\
Shivchander Sudalairaj$^*$ \\
Abhishek Bhandwaldar$^*$ \\
Aldo Pareja$^*$ \\
Kai Xu \\
David D. Cox \\
Akash Srivastava$^{*,\dagger}$ \\
\\
*Equal Contribution, $\dagger$Corresponding Author
}

\newcommand{\tabref}[1]{table~\ref{#1}}

\newcommand{\labl}{Labradorite-13b}
\newcommand{\labm}{Merlinite-7B}

\iclrfinalcopy % Uncomment for camera-ready version, but NOT for submission.
\begin{document}

\maketitle

\begin{abstract}
This work introduces LAB (Large-scale Alignment for chatBots), a novel methodology designed to overcome the scalability challenges in the instruction-tuning phase of large language model (LLM) training. Leveraging a taxonomy-guided synthetic data generation process and a multi-phase tuning framework, LAB significantly reduces reliance on expensive human annotations and proprietary models like GPT-4. We demonstrate that LAB-trained models can achieve competitive performance across several benchmarks compared to models trained with traditional human-annotated or GPT-4 generated synthetic data. Thus offering a scalable, cost-effective solution for enhancing LLM capabilities and instruction-following behaviors without the drawbacks of catastrophic forgetting, marking a step forward in the efficient training of LLMs for a wide range of applications.
% We challenge the common way of training chatbots that has a heavy pertaining stage followed by a light fine-tuning/alignment stage.
% We introduce LAB, a new alignment method for chatbots that performs supervised fine-tuning using orders of magnitude more data than standard alignment approaches.
% To obtain a diverse and targeted dataset, a taxonomy-driven synthetic data generation method is developed to generate data covering knowledge, foundational skills and compositional skills over a wide range of tasks.
% The taxonomy is not only used for data curation but also drives the sampling of seed examples in data generation process to ensure the generated data is diverse and high quality.
% In order to learn both knowledge and skills, we propose a phased-training approach with replay buffers in two phases, knowledge tuning and skills tuning, where the replay buffer ensures that all data can be properly learned by the model.
% By applying LAB to align \textsc{Llama-2-13b} and \textsc{
% Mistral-7B} with \textsc{Mixtral-8x7B} as a teacher model,
% we obtain the best fine-tuned models in terms of MT-Bench on the LMSYS Chatbot Arena Leaderboard while other metrics are also competitive.\\
% We open-source the weights of our aligned models, available at \url{https://}.
\end{abstract}

\section{Introduction}

Large language models (LLMs) have achieved remarkable levels of success in various natural language processing (NLP) applications, including question-answering \cite{}, entity extraction \cite{}, and summarization \cite{}. This has been made possible, in large part, by the introduction of the transformer architecture \cite{}, which can leverage large amounts of unlabeled, unstructured data, enabling the scaling of LLMs to billions, or even trillions of parameters. LLMs are typically trained in phases: a self-supervised pre-training phase, followed by supervised alignment tuning phases. 

The majority of the cost of training an LLM comes from the pre-training phase. During this phase, a model is trained in an auto-regressive manner to predict the next token in the target language using trillions of tokens worth of unlabeled data, requiring thousands of GPUs training for months at a time. Alignment tuning, typically happens in two stages: instruction tuning, followed by preference tuning. Instruction tuning is more akin to the traditional model training approach in machine learning, where the model is trained directly on tasks of interest. In this stage, the model is given a task description in the form of an natural language instuction (e.g. \textit{Summarize the following news article in 2 lines: \{News article\}}) and the model is trained to maximize the likelihood of the provided ground truth summary. Preference tuning, on the other hand, is done using techniques such as RLHF \citep{stiennon2022learning,ouyang2022training} and DPO \citep{rafailov2023direct}, where the response from an instruction-tuned model is rated as preferred or unpreferred using human feedback. 

In comparison to pre-training, the instruction tuning and preference tuning stages comprise a small fraction of the overall training procedure, both in terms of the data used as well as the compute infrastructure required to train models \cite{touvron2023llama}. For example, Meta's LLaMA 2 models were trained with just tens of thousands of high quality human-generated instruction/response data pairs, followed by multiple rounds of RLHF with a comparatively limited number of examples as compared to pretraining data volumes \cite{touvron2023llama}. From a traditional machine learning training perspective, this imbalance in the scale across the phases is unconventional---typically one would expect a model to perform best when it has been trained directly on the desired tasks, using as much data as possible. The deviation from the tradtional LLM approach relies on the idea that pretraining captures enough of the distribution of language and knowledge, such that a small amount of supervised training can ``unlock'' or shape latent abilities related to the ultimate desired instruction-following behavior of the model. However, unlike the unstructured data that is abundantly available in the public domain, high-quality, human-generated task-specific instruction data is costly to procure, even via crowd-sourcing, and human-generated instruction data is typically closely guarded by model builders, even for ostensibly ``open'' model-building efforts.
In this work, we address the challenges associated with scaling of the alignment-tuning phase and propose a new method called LAB: Large-scale Alignment for chatBots. % DDC: I don't think it's actually necessary to use the "labrador", as opposed to LAB, name here (here affectionately referred to as ``Labrador'').
The LAB method consists of two components: (i) a taxonomy-guided synthetic data generation method and quality assurance process that yields a highly diverse and high-quality instruction dataset, without resorting to the use of proprietary LLMs like GPT-4 or substantial human curation, and (ii) a novel multi-phase training framework and unconventional tuning regime that allows for adding new knowledge and instruction-following abilities into pre-trained LLMs without suffering from catastrophic forgetting. Our findings show that LAB-trained models can perform competitively with proprietary and open-source models that use human annotations and/or synthetic data generated using GPT-4 on a number of benchmarks. 

\section{Related Work}

Existing methods for instruction tuning typically either rely on humans for generating high-quality datasets, or use synthetic data generation using a large teacher model.  
%GPT-3.5/4, Claude, and LLaMA 2 rely on human-annotated data for supervised fine-tuning (SFT) and reinforcement Learning from Human Feedback (RLHF). 
OpenAI \citep{ouyang2022training} arguably set the standard for model alignment from human data, employing human annotators to gather data for supervised fine tuning (SFT) and reinforcement learning with human feedback (RLHF) training. Collecting human-generated data for these steps is complex undertaking; the selection of annotators requires a rigorous multi-stage screening process aimed at achieving high inter-annotator agreement, and collecting even modest amounts data (by LLM standards) requires the coordination of large groups of annotators.
The creators of the LLaMA 2 model series \citep{touvron2023llama}  followed a similar recipe, collecting tens of thousands of human-generated instruction samples, and approximately 1 million human-annotated binary comparisons for reward modeling. 
Not only are such approaches expensive and time consuming, but they can also potentially limit agility in exploring the space of instructions and capabilities the model is trained to perform.
%Such an extensive undertaking significantly constrains the incorporation of new skills while striving to maintain consistency with the preferences reflected in previously acquired data.
Alternatives to this approach, such as transforming existing human datasets into instructions via templating \citep{weifinetuned} can be more cost effective, but face limitations in the naturalness and length of the responses used for training.

More recently, training with synthetic data generated from LLMs has emerged as an alternative to purely human-data-based approaches.
\cite{wang2023selfinstructaligning} introduced Self-Instruct, which leverages a small number of handwritten human seed instructions as input to bootstrapping process to generate a large number of samples using an LLM's own generation abilities.
\cite{alpaca} built upon Self-Instruct, using a larger teacher model to generate synthetic data to train a smaller student model, and incorporating principles in the generation prompt to promote diversity in the generated instruction data.
\cite{xu2023wizardlmempowering} introduces Evol-Instruct, another variant of Self-Instruct, that synthesizes iteratively more complex instruction to overcome shortcomings of previous methods. 
\cite{mukherjee2023orca}, \cite{mitra2023orcateaching} present a synthetic data generation approach to enhance task diversity and scalability, alongside a progressive training framework aimed at improving the model's reasoning ability and response style to match teacher models. This is achieved by generating rich reasoning signals in the generated answer and progressively training on datasets of varying difficulty in incremental phases. 

Similar to LAB, concurrent work, GLAN \citep{li2024synthetic}, employs a semi-automatic approach to synthetic data generation that uses a human-curated taxonomy to generate instruction tuning data from a teacher model. However, as explained in section \ref{sec:know-sdg}, unlike LAB, GLAN cannot be used to generate synthetic data from domains that are not captured in the teacher model's support. As such, while LAB uses the open-source Mixtral model as the teacher, like many other synthetic data generation approaches, GLAN has to rely on a large proprietary model (GPT-4). This poses complicated questions about the usability of generated data (especially for commercial purposes) since the terms of use of proprietary models typically forbid using the model to improve other models.

\section{Methodology}

\begin{figure}
\hspace*{-5.5em}
    \centering
\begin{tikzpicture}[level 1/.style={sibling distance=12em},
                    level 2/.style={sibling distance=5em},
                    level distance=3em]
\tikzstyle{ex} = [rectangle, draw, rounded corners, fill=blue!20]
\tikzstyle{data} = [cylinder, shape border rotate=90, aspect=0.05, draw, fill=gray!20]
\tikzstyle{doc} = [rectangle, draw, rounded corners, fill=gray!20]
% database/.style={
%   cylinder,
%   cylinder uses custom fill,
%   cylinder body fill=yellow!50,
%   cylinder end fill=yellow!50,
%   shape border rotate=90,
%   aspect=0.25,
%   draw
% }

\node {taxonomy root}
    child {node {knowledge}
      child {node {$\dots$}}
      child {node {textbook}
        child [level distance=3em] {node (finance) {finance}
          child {node [ex] (knowledge-ex-1) {example 1}}
          child {node [ex] (knowledge-ex-2) {example 2}}
          child {node [ex] (knowledge-ex-3) {example 3}}
        }
      }
      child {node {$\dots$}}
    }
    child {node [align=center] {foundational\\skills}
      child {node {$\dots$}}
      child {node {mathematics}
        child [level distance=2em] {node {arithmetic}
          child {node {addition}
            child [level distance=2.5em] {node [data] (cot-data) {math instruct data}}
          }
        }
      }
      child {node {$\dots$}}
    }
    child {node [align=center] (c-skills) {compositional\\skills}
      child {node {$\dots$}}
      child {node {writing}
        child [level distance=2em] {node {email}
          child {node {earnings report}
            child {node [ex] (c-skills-ex-1) {example 1}}
            child {node [ex] (c-skills-ex-2) {example 2}}
            child {node [ex] (c-skills-ex-3) {example 3}}
          }
        }
      }
      child {node {$\dots$}}
    };

\node (document) [doc, left of=knowledge-ex-1, xshift=-2.5em] {document};
\node (anchor) [below of=cot-data, yshift=-1em] {};
\node (sdg-1) [shape=rectangle, draw, below of=knowledge-ex-2, yshift=-1em, text width=14em, text centered] {Synthetic Data Generator 1};
\node (sdg-2) [shape=rectangle, draw, below of=c-skills-ex-2, yshift=-1em, text width=14em, text centered] {Synthetic Data Generator 2};

\node [data, below of=sdg-1, yshift=-0.5em] (synth-art-history-data) {synthetic ``finance'' data};
\node [right of=synth-art-history-data, xshift=4.5em] {0.1--2k};
\node [data, below of=sdg-2, yshift=-0.5em] (synth-outline-data) {synthetic ``email'' data};
\node [right of=synth-outline-data, xshift=3.9em] {0.1--2k};

\node [below of=anchor, shape=rectangle, draw, text width=32em, text centered, yshift=-4em, minimum height=2em] (phased-training) {Phased training};
\node [left of=phased-training, xshift=-20em] (llm) {pre-trained LLM};
\draw[-] (llm) -- (phased-training);

\draw[-] (finance) -| (document);
\draw[->] (document) |- (sdg-1);
\draw[->] (knowledge-ex-1) -- (sdg-1);
\draw[->] (knowledge-ex-2) -- (sdg-1);
\draw[->] (knowledge-ex-3) -- (sdg-1);
\draw[->] (sdg-1) -- (synth-art-history-data);
\draw[->] (synth-art-history-data) -- (phased-training);

\draw[->] (cot-data) -- (phased-training);

\draw[->] (c-skills-ex-1) -- (sdg-2);
\draw[->] (c-skills-ex-2) -- (sdg-2);
\draw[->] (c-skills-ex-3) -- (sdg-2);
\draw[->] (sdg-2) -- (synth-outline-data);
\draw[->] (synth-outline-data) -- (phased-training);

\node (anchor-sec-1) [above right of=c-skills, xshift=5em, yshift=2em] {};
\node (anchor-sec-2) [below right of=c-skills-ex-2, xshift=5em] {};
\node (anchor-sec-3) [below of=anchor-sec-2, yshift=-4em] {};
\node (anchor-sec-4) [below of=anchor-sec-3, yshift=-1em] {};

\draw [decorate, decoration={brace, amplitude=5pt, mirror, raise=4ex}] (anchor-sec-2) -- (anchor-sec-1) node [midway,xshift=5em]{\secref{sec:tax}};

\draw [decorate, decoration={brace, amplitude=5pt, mirror, raise=4ex}] (anchor-sec-3) -- (anchor-sec-2) node [midway,xshift=5em]{\secref{sec:sdg}};

\draw [decorate, decoration={brace, amplitude=5pt, mirror, raise=4ex}] (anchor-sec-4) -- (anchor-sec-3) node [midway,xshift=5em]{\secref{sec:train}};

\end{tikzpicture}
    \caption{Overview of the LAB alignment method. Starting from the taxonomy root, data are curated in each top-level groups and examples in the leaf nodes are used by the synthetic data generators to generate orders of magnitude data for the phased-training step for instruct-tuning.}\label{fig:overview}
\end{figure}
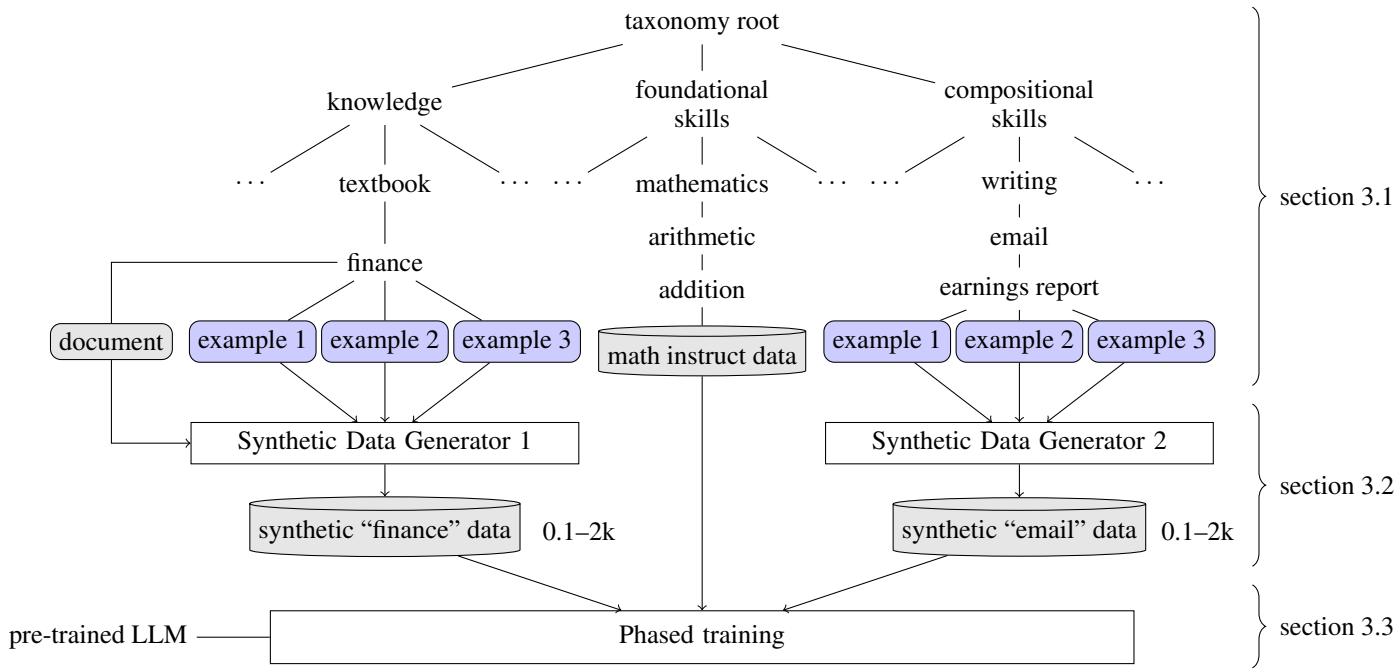

LAB consists of two components: (i) a taxonomy to enable data curation (\secref{sec:tax}) as well as, guide the synthetic data generator (\secref{sec:sdg}) and (ii) a multi-phased instruction-tuning method with replay buffers to enable large-scale alignment-tuning. (\secref{sec:train}).
(i) serves the purpose of ensuring high diversity and quality in the synthetically generated instruction-tuning dataset while (ii) ensures training stability and prevents catastrophic forgetting. 
\Figref{fig:overview} provides an overview of the end-to-end pipeline of applying the LAB method to align a pre-trained LLM.

% \Figref{fig:pipe} provides an overview of the end-to-end pipeline with these three components of applying the Labrador method to a pre-trained LLM.
% \begin{figure}[t]
% \begin{center}
% \includegraphics[width=0.8\textwidth]{figs/pipe.png}
% \end{center}
% \caption{Full Labrador pipeline. Taxonomy is used by the synthetic data generator (SDG) to create large amounts of knowledge and skill data. This data is then filtered using the same teacher model that generated the data. The filtered data is used in two phases to train the target pre-trained LLM.\label{fig:pipe}}
% \end{figure}

\subsection{Taxonomy}\label{sec:tax}

To enable the data curator or the model designer to organize the instruction-tuning training data, we define a taxonomy that hierarchically classifies the data samples into smaller task groups. At a high level, the taxonomy has three main branches: knowledge, foundational skills, and compositional skills. Each of these branches is further split into more granular levels where the tasks are defined in the leaf nodes and exemplified by providing manually written instruction-response pairs. This allows for easily identifying missing tasks in the target LLM and other tasks of interest and adding them to the training data pool. New tasks are added to the taxonomy by creating a leaf node under the appropriate branch and attaching 1--3 examples.

\paragraph{Knowledge}
The knowledge branch in the taxonomy is first divided based on document types like textbooks, technical manuals, etc., which are further divided into various domains like finance, statistics, etc.; see the sub-tree for knowledge in \Figref{fig:overview} as an example. 
Each domain has a collection of documents and a sample set of domain-specific questions and answers.
This organization allows for better control over the licensing of text documents. As described in the next section, only the documents with permissible licenses are selected for synthetic data generation, excluding knowledge sources that lack proper licensing, reinforcing the integrity of our knowledge-generation processes. 
% We selectively employ text documents possessing over the permissible licenses for data generation, thereby ensuring adherence to legal and ethical standards. Moreover, this approach enables us to exclude knowledge lacking proper licensing, further reinforcing the integrity of our knowledge-generation processes. 

\paragraph{Foundational skills}
We identify mathematics, coding, linguistic ability and reasoning as foundational skills that the model requires to prime itself for better knowledge acquisition and build further complex and compositional skills.
To teach the model foundational skills, we employ publicly available datasets \citep{flan,mathins,conala,musique}; see the sub-tree for foundational skills in \Figref{fig:overview} for an example.

\paragraph{Compositional skills}
Compositional skills refer to the tasks that require a combination of knowledge and foundational skills, synergistically, to answer complex queries from users. For instance, the model's ability to write a company-wide email sharing insights about the company's performance last quarter and guidance for the upcoming year would require the model to understand the financial aspects of revenue, profit and loss, the skills of doing basic arithmetic and also have the skills to compose a formal email.

% Our approach begins by crafting a taxonomy that encompasses knowledge, foundational skills, and compositional skills, as exemplified in \figref{fig:tax}.
% \begin{figure}[t]
% \begin{center}
% \includegraphics[width=0.8\textwidth]{figs/tax-full.png}
% \end{center}
% \caption{Partial taxonomy of the training data used for {\labl} and {\labm} models. It is divided into three buckets of knowledge, foundational and composition skills }\label{fig:tax}
% \end{figure}
% This taxonomy is designed to have a fine-grained level at each domain, ensuring comprehensive coverage in the teacher representation space. 
% In each leaf node of the taxonomy, we strategically incorporate 3-5 seed examples. Subsequently, we explicitly sample additional instruction data pairs from the teacher model for each specific leaf node. This meticulous process guarantees that the generated synthetic data exhibits diversity equivalent to the knowledge space represented by the taxonomy.

% Furthermore, the taxonomy-driven synthetic data generation approach not only addresses the limitations associated with existing methodologies but also provides a way to go beyond the limits of the teacher model's knowledge. The carefully curated taxonomy serves as a structured framework that not only captures the existing knowledge and skills of the teacher model but also offers further expansion.

\subsection{Taxonomy-driven Synthetic Data Generator}\label{sec:sdg}

The small number of manually curated data samples, embedded in the leaf nodes of the taxonomy, can be directly used for instruction tuning of the chatbot, however, the model may still perform poorly. Prior work \citep{humpback} has shown that typically, a large amount of high-quality instruction data is required for improving instruction following performance of LLMs. It is possible to leverage existing SDGs like \cite{wang2023selfinstructaligning, alpaca} to use the embedded examples and generate a lot more instruction data synthetically using teacher LLMs. But, such distillation-based SDGs tend to over-sample from the dominant modes of the teacher model and thus lack in diversity and quality of the generated data \cite{gudibande2023false}. 
We argue that this limitation is attributed to the random selection of examples from the pool of seed samples: with random selection, the examples used to prompt the teacher model at each time are an ``average'' of the seed pool i.e. they do not \emph{focus} on any specific task. This lack of focus tends to encourage the teacher model to generate more synthetic data from its dominant modes and ignore the long tail of interesting tasks.

\begin{figure}[h]
\begin{center}
\begin{subfigure}[b]{0.48\textwidth}
    \centering
     \includegraphics[width=\textwidth]{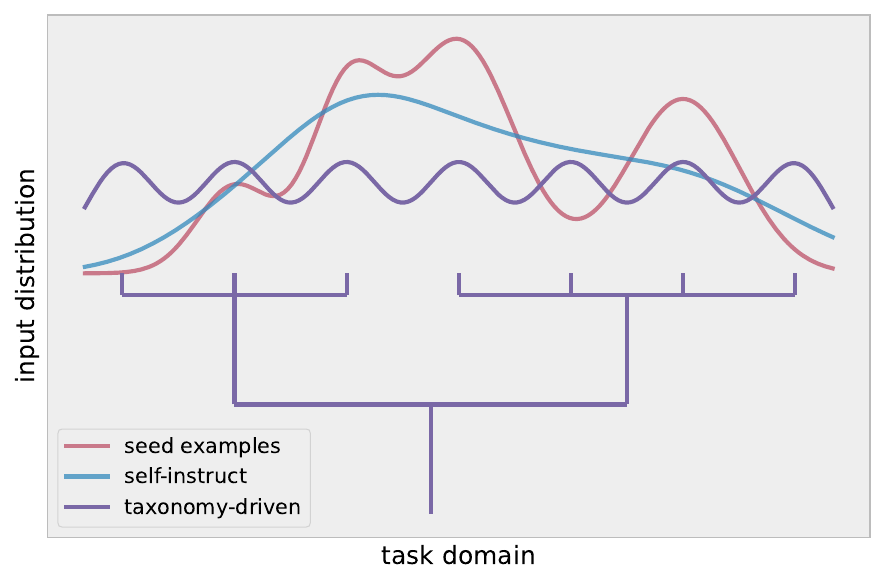}
     \caption{Input distributions}\label{fig:intuition-input}
\end{subfigure}
\begin{subfigure}[b]{0.48\textwidth}
    \centering
     \includegraphics[width=\textwidth]{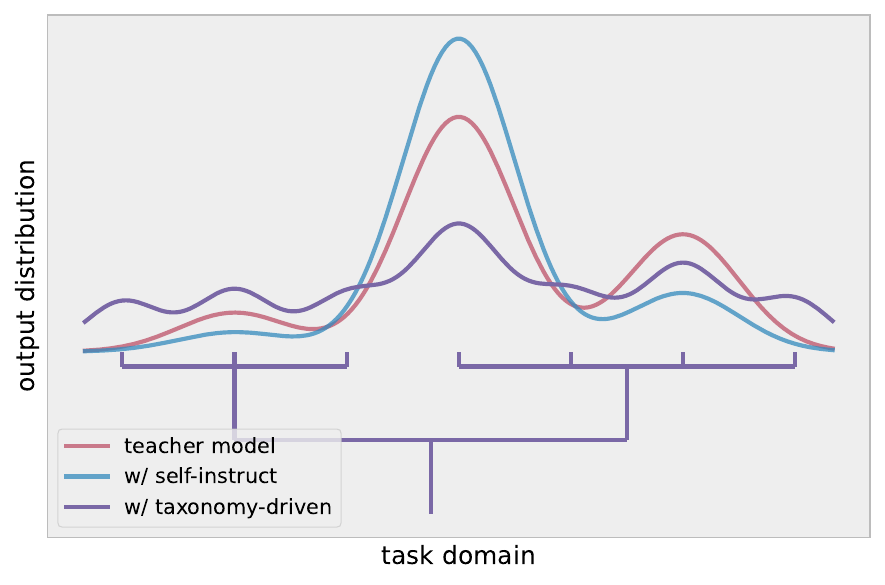}
     \caption{Output distributions}\label{fig:intuition-output}
\end{subfigure}
\end{center}
\caption{Intuition of how taxonomy-driven sampling produces diverse set of synthetic data and hence improve the data used to train student model across the task domain. \Figref{fig:intuition-input} shows how taxonomy-driven sampling leads to an input distribution with wide support and distinct modes while self-instruct gives an smooth input distribution. \Figref{fig:intuition-output} shows the consequence using inputs in generating synthetic data: teacher model will focus its own dominant modes if the input is smooth but focus on each task better if the inputs are also concentrated on each task.}\label{fig:intuition}
\end{figure}

To address this issue, we replace the random sampling in existing SDGs with a taxonomy-driven approach to guide the sampling of synthetic data, enabling targeted coverage of the support of the teacher model distribution around the individual leaf nodes of the taxonomy.
% Essentially, the diversity in the taxonomy enables the teacher model to explore a wide range of tasks and the taxonomy-driven sampling encourages the teacher model to exploit each task extensively, i.e.~targeted exploration of the entire taxonomy support.
\Figref{fig:intuition} illustrate the high-level idea behind this change.
\Figref{fig:intuition-input} shows the issue of randomly sampling in the \emph{input} space of the teacher model (i.e.~prompts).
Given a set of seed examples (red), randomly sampling with more than one example gives an approximation to the average of the seed pool, leading to a smoothed distribution, e.g.~ self-instruct distribution (blue).
With the taxonomy-driven sampling, since only the examples within each of the leaf nodes are used when sampling for the corresponding tasks, 
% each of the tasks are guaranteed to be captured in the synthetic data distribution (purple).
each of the tasks are guaranteed to be well represented in the prompts (purple).
Second, when it comes to the \emph{output} space, for a given teacher model (red), prompting it with random examples (i.e.~smoothened input distribution) tends to make it sampling from its own dominant mode (blue) while prompting it with focused examples in each leaf node (i.e.~input distribution with distinct modes), 
the teacher model is guaranteed to generate synthetic data for each of the tasks (purple).

With the above insight, we now introduce two new synthetic data generation (SDG) methods in LAB that leverage the taxonomy to guide the data generation process. The first one is targeted for skills generation and uses the handful of task examples in the leaf nodes to generate a lot more using the open-source Mixtral-7x8B model.  The second one is targeted at knowledge generation. While it still uses the Mixtral-7x8B model, unlike prior works, it does not rely on the knowledge stored in the teacher model.

% \begin{figure}[h]
% \begin{center}
% \includegraphics[width=0.8\textwidth]{figs/datagen.png}
% \end{center}
% \caption{Overview of synthetic data generation pipeline}\label{fig:datagen}
% \end{figure}

\begin{figure}[]
    \centering
\begin{tcolorbox}
\small

You are asked to come up with a set of \{num\_samples\} diverse questions on \{task\}.\\

Please follow these guiding principles when generating responses:\\

* Use proper grammar and punctuation.\\
* Always generate safe and respectful content. Do not generate content that is harmful, abusive, or offensive.\\
* Always generate content that is factually accurate and relevant to the prompt.\\
* The questions should be clear and human-like.\\
* The questions should be diverse and cover a wide range of topics.\\
* The questions should not be template-based or generic, it should be very diverse.\\
* Simply return the questions, do not return any answers or explanations.\\
* Strictly adhere to the prompt and generate responses in the same style and format as the example.\\

To better assist you with this task, here is an example:\\
\#\#\# Question:\\
1. \{icl\_question\}\\

Now generate \{num\_samples\} such questions, remember to follow the principles mentioned above and use the same format as the examples. Remember to use the same style and format as the example above. Return your responses in the format of [\#\#\# Question [question number]: [question]]
\end{tcolorbox}
\caption{Instruction Generator prompt template}\label{fig:question-template}
\end{figure}
\subsubsection{Skill Generation}
Skills-SDG uses four prompt templates, one for each of the four, below-mentioned, stages of data generation. Each template has its own set of principles and instructions that control the role of the teacher model (generator vs evaluator) and guide the generation/evaluation process.

\begin{figure}[]
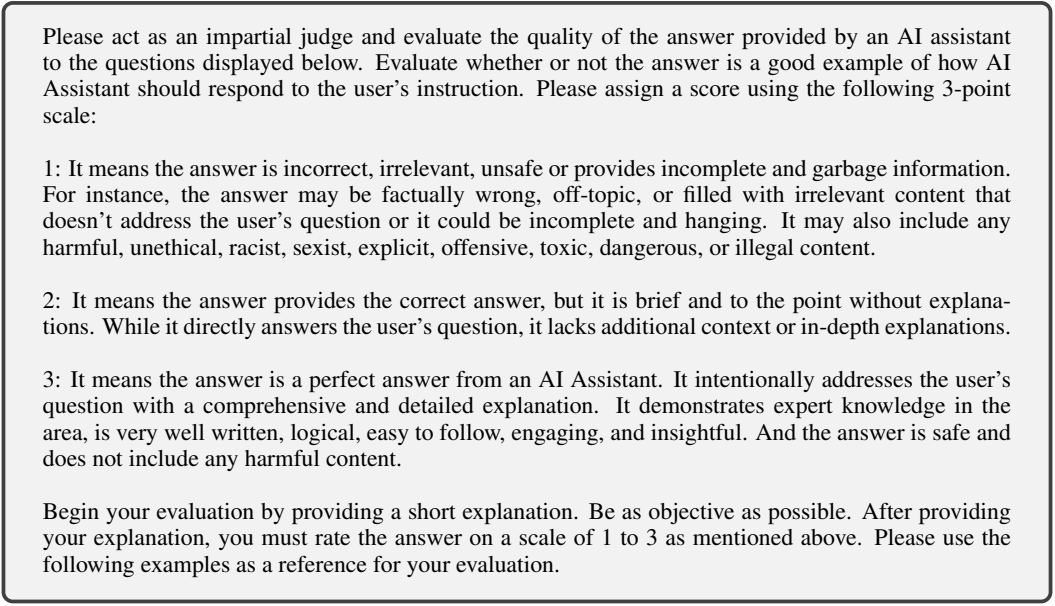

    \centering
\begin{tcolorbox}
\small

Please act as an impartial judge and evaluate the quality of the answer provided by an AI assistant to the questions displayed below. Evaluate whether or not the answer is a good example of how AI Assistant should respond to the user's instruction. Please assign a score using the following 3-point scale:\\

1: It means the answer is incorrect, irrelevant, unsafe or provides incomplete and garbage information. For instance, the answer may be factually wrong, off-topic, or filled with irrelevant content that doesn't address the user's question or it could be incomplete and hanging. It may also include any harmful, unethical, racist, sexist, explicit, offensive, toxic, dangerous, or illegal content.  \\

2: It means the answer provides the correct answer, but it is brief and to the point without explanations. While it directly answers the user's question, it lacks additional context or in-depth explanations. \\

3: It means the answer is a perfect answer from an AI Assistant. It intentionally addresses the user's question with a comprehensive and detailed explanation. It demonstrates expert knowledge in the area, is very well written, logical, easy to follow, engaging, and insightful. And the answer is safe and does not include any harmful content. \\

Begin your evaluation by providing a short explanation. Be as objective as possible. After providing your explanation, you must rate the answer on a scale of 1 to 3 as mentioned above. Please use the following examples as a reference for your evaluation.

\end{tcolorbox}
\caption{Instruction-response Evaluation template}\label{fig:qaevaluation-template}
\end{figure}

\begin{enumerate}
    % \item Generating teacher prompts: Task specific instructions and generation principles are combined with the examples
    \item \textbf{Instruction generation:} In the first stage, the teacher model acts as a question generator, using a specialized prompt (see \Figref{fig:question-template} for an example) to leverage its knowledge and create diverse questions. By iterating through each leaf node of a taxonomy, the teacher generates queries that adhere to specific principles and thoroughly explore the targeted domain, enhancing the comprehensiveness of the generated content. 
    \item \textbf{Evaluating synthetic instruction:} In this stage, the teacher model assumes the role of an instruction evaluator, the teacher model uses targeted prompts to filter out questions that don't meet predefined principles, including relevance to the domain, potential harm, or questions beyond a language model's answering capabilities. This ensures that only high-quality, contextually appropriate questions move forward in the process.
    \item \textbf{Generating responses:} The teacher model, functioning as a response generator in this stage, adopts dual personas for precision and creativity, guided by distinct prompts. This tailored approach helps to generate both, creative responses for domains like writing and role-play, and precise answers for STEM and data extraction, aligning the response style to human expectations through principles and seed examples in the leaf nodes.
    \item \textbf{Evaluating the synthetic instruction-response pair:} The final stage involves a rigorous process to filter and select high-quality instruction and response pairs. Using a 3-point rating system (see \Figref{fig:qaevaluation-template} for an example), the teacher model evaluates each sample, filtering out those that are incorrect, irrelevant, or deviate from the provided principles, ensuring the training dataset's quality and relevance are enhanced for the student model.

\end{enumerate}

% \paragraph{Question-Answer Evaluation}
% Building on insights from self-curation of instruction data, as demonstrated in prior research such as \cite{humpback}, and leveraging language models' evaluative capabilities showcased by works like \cite{mtbench}, our pipeline incorporates a rigorous process for filtering and selecting high-quality question and answer pairs. In this stage, the teacher model assumes the role of an evaluator (See \figref{fig:qaevaluation-template}), discerning the merit of each sample generated during the previous phases.

% Utilizing a rating system, samples are systematically assessed on a 3-point scale. This evaluative process aims to filter out incorrect, irrelevant, or any samples that deviate from the generation principles established earlier. This approach ensures that only the most pertinent and principled question and answer pairs progress to the subsequent stages of the training pipeline. By integrating this evaluation step, we refine the training dataset, enhancing the quality and relevance of the instruction data for the student model.
\subsubsection{Knowledge-generation}
\label{sec:know-sdg}
Synthetic data generators are inherently limited by the knowledge and capabilities of the teacher model. This is one of the main reasons why most successful SDG methods \citep{xu2023wizardlmempowering, mukherjee2023orca, mitra2023orcateaching} depend on GPT-4 model, which presumably has the highest coverage of knowledge and skills. However, there are many domains that no open/proprietary model is trained on and hence cannot work as a teacher model using existing SDG methods. To address this limitation, in LAB we devised a new SDG pipeline for generating instruction data on domains that the teacher model has not been trained on. We call it knowledge-SDG.

Similar to the process of skills generation, knowledge-SDG uses the curator-provided examples embedded in the leaf nodes of the knowledge branch of the taxonomy. But additionally, the teacher model is provided a knowledge source in the form of documents, manuals, and books on the target subject to ground the generated instruction data into a reliable source thus avoiding dependence on the internal knowledge base of a teacher model, which may struggle with specialized domains and could lead to inaccuracies or hallucinations especially on highly specialized, technical domains. 

% \paragraph{Prompt Crafting}
% Knowledge generation involves crafting a prompt that delineates instructions for formulating textbook-style questions and answers, alongside selecting an appropriate document from which these questions and answers are generated. The generation starts by taking examples in each leaf node of the knowledge taxonomy, crafting prompts for all the documents with the domain-specific seed examples, and prompting a large teacher model to generate more questions and answers from the document.

% By anchoring the teacher model's responses in documents, we not only uphold knowledge groundedness but also facilitate diversity in prompts and responses. For instance, while generating comparative-type questions and answers, we may include examples involving the comparison of two products. However, even if the input document focuses on a single product or lacks mention of any products, the model can still generalize to comparisons based on diverse product attributes or extend its capabilities to other categories, such as methodologies in research papers or conceptual comparisons in textbooks.

% We also generate instruction data that closely resembles natural instructions by pairing documents from a specific domain with questions posed by individuals from a different domain. For instance, we might pair questions and answers typically posed by a lawyer with a mathematical research article.

% \paragraph{Generated Knowledge Verification}
To ensure that the generated answers remain faithful to the content of the source material, similar to the skills-SDG, teacher model is repurposed as an evaluator that validates the generated responses are grounded and faithful to the source documents.

\subsection{Multi-Phase Training}\label{sec:train}

LAB training happens in two phases, knowledge tuning, followed by skills tuning. 

In the knowledge-tuning phase, the model is trained on samples from the knowledge and foundational skills branches of the taxonomy. This phase in-turn, is carried out in two steps. We split the data under the knowledge and foundational skills branches into two buckets based on the response length. Then we first train the model on the samples with short responses before moving on to training on samples with long responses. Similar to prior work \citep{mitra2023orcateaching}, our empirical results also suggest that this two-step approach to knowledge-tuning improves model performance.

Post-knowledge tuning, we start the skills-tuning phase where the best model checkpoint from the knowledge-tuning phase is trained on the compositional skills branch of the taxonomy. To address the challenge of catastrophic forgetting when training in two distinct phases, a replay buffer of the data from the knowledge-tuning phase in employed. Our empirical findings indicate that starting with knowledge and foundational skills training, before progressing to compositional skills leads to significantly better benchmark performance.

For selecting the best model checkpoint during intermediate phases, we rely on the MMLU benchmark \citep{hendrycks2020measuring} during the knowledge-tuning phase and the MT-bench \citep{zheng2024judging} during the skills-tuning phase. Please refer to \tabref{tab:phased-training} for an overview of our training phases.

\paragraph{Training Details}
In our training process, we consciously avoid overtraining. Despite the possibility of achieving higher scores on intermediate benchmarks, we have found that selecting checkpoints from earlier stages of training results in more reliable and generalizable model performance. We employ small learning rates with an extended warm-up period, specifically $2 \times 10^{-5}$ for Llama-based models and $1 \times 10^{-6}$ for Mistral-based models, each beginning with a linear warm-up. This strategy is hypothesized to aid the model in transitioning from broad dataset-wide learning to more focused, task-specific adjustments. Additionally, we utilize a large effective batch size of 3840, achieved through gradient accumulation, to enhance stability across the diverse range of tasks being learned concurrently. Our findings suggest that using cosine decay on learning rates during intermediate phases can destabilize subsequent training stages, likely due to the learning rate's reduction to near zero, narrowing the loss landscape and complicating the integration of new phase gradients. Refer to \tabref{tab:hps} for an overview of our training hyper-parameters.

% \paragraph{Incremental Knowledge and Skills Inclusion}
% Our phased training strategy, supported by a taxonomy-driven approach, allows for the systematic inclusion of new knowledge and skills. Which at the same time could accommodate new benchmarks that test these additions. This structured progression guarantees the model's adaptability and continued growth in performance across diverse domains.

% \begin{figure}[h]
%     \begin{center}
%     \includegraphics[width=0.6\textwidth]{figs/training.png}
%     \end{center}
%     \caption{}\label{fig:phased-training}
% \end{figure}

\begin{table}[]
\centering
\begin{tabular}{lllll}
\toprule
Phase            & Step & Training data                                                                  & Replay buffer &  \\
\hline
\multirow{2}{*}{Knowledge Tuning} & 1    & Knowledge (short)                                                               &             N/A  &  \\\cline{2-5}
 & 2    & \makecell[l]{Knowledge (long)\\Foundational skills} &       KT/1 data       &  \\\hline
Skill Tuning     &  N/A    & Compositional skills   & KT/1 \& KT/2 data              & \\ \bottomrule
\end{tabular}
\caption{Data and reply buffers used in phase-training.}\label{tab:phased-training}
\end{table}

\section{Results}

In this study, we implemented the LAB method on two distinct open models, \textsc{Llama-2-13b} \citep{touvron2023llama}and \textsc{Mistral-7B} \citep{jiang2023mistral7b}, utilizing \textsc{Mixtral-8x7B-Instruct-v0.1} \citep{jiang2024mixtralexperts} as the teacher model. This approach yielded two LAB-aligned models: \textsc{\labl} and \textsc{\labm}.

During the synthetic data generation phase, we employed a taxonomy consisting of numerous leaf nodes to produce a dataset comprising 1.2 million samples, divided almost evenly between knowledge-based (617k) and skill-based (588k) samples. The specific training hyper-parameters employed during this study are summarized in \tabref{tab:hps}.

\begin{table}[t]
% \hspace*{-4em}
    \setlength{\tabcolsep}{2pt}
    \centering
    \begin{small}\begin{sc}
    \begin{tabular}{ll|ccccccc}
    \toprule
    Model   & Phase/step          & \makecell{Learning\\rate} & \makecell{Batch\\size} & \makecell{Context\\length} & \#samples & \#warm-up   &  \#epochs    \\ \hline
    \multirow{3}{*}{\labl}  & KT/1 & \multirow{3}{*}{2e-5} & \multirow{3}{*}{3840} & \multirow{2}{*}{2048} & 630k & \multirow{3}{*}{385}&  5\\ 
                            & KT/2 &                       &                       &                       & 230k &                   & 7\\ 
                            & ST   &                       &                       & 4096                  & 550k &                   & 7\\ \hline
    \multirow{3}{*}{\labm} & KT/1  & \multirow{3}{*}{1e-6} & \multirow{3}{*}{3840} & \multirow{2}{*}{2048} & 630k & \multirow{3}{*}{800}& 4\\ 
                           & KT/2  &                       &                       &                       & 230k &                   & 4\\ 
                           & ST    &                       &                       & 4096                  & 550k &                   & 7\\ 
    \bottomrule
    \end{tabular}
    \end{sc}\end{small}
    \caption{Hyper-parameters used in training for \textsc{\labl} and \textsc{\labm}.}\label{tab:hps}
    \setlength{\tabcolsep}{6pt}
\end{table}

We compare the performance of \textsc{\labl} and \textsc{\labm} against other models that use the same base models for alignment, which include
\paragraph{\textsc{Llama-2-13b}}
\begin{itemize}
    \item \textsc{Llama-2-13b-chat} \citep{touvron2023llama}: RLHF with human annotators by the same team that develops \textsc{Llama-2-13b}
    \item \textsc{Orca-2} \citep{mitra2023orcateaching}: 
    \item \textsc{WizardLM-13B-V1.2} \citep{xu2023wizardlmempowering}: model with the highest MT-Bench amongs those use \textsc{Llama-2-13b} as the base model on LMSYS Chatbot Arena Leaderboard \citep{mtbench}.
\end{itemize}
\paragraph{\textsc{Mistral-7B}}
\begin{itemize}
    \item \textsc{Mistral-7B-Instruct-v0.2} \citep{jiang2023mistral7b}: instruction-tuning using supervised fine-tuning (SFT) on publicly available conversation datasets by the same team that develops \textsc{Mistral-7B}
    \item \textsc{Zephyr-7b-beta} \citep{tunstall2023zephyrdirect}: model with the highest MT-Bench amongs those use \textsc{Mistral-7B} as the base model on LMSYS Chatbot Arena Leaderboard \citep{mtbench}. 
\end{itemize}
To compare the aligned LLMs, we consider the following evaluation metrics with the settings consistent with those used by LMSYS Chatbot Arena Leaderboard \citep{mtbench}
\begin{itemize}
    \item MT-Bench \citep{mtbench}: 1-turn and 2-turn average
    \item MMLU \citep{mmlu}: 5-shot
    \item ARC \citep{clark2018thinkyou}: 25-shot
    \item HellaSwag \citep{zellers2019hellaswagcan}: 10-shot
    \item Winogrande \citep{sakaguchi2019winograndeadversarial}: 5-shot
    \item GSM8k \citep{cobbe2021trainingverifiers}: 5-shot strict 
\end{itemize}
All results are reported in \tabref{tab:res}.
% \begin{landscape}
\begin{table}[t]
\hspace*{-5em}
    \setlength{\tabcolsep}{2pt}
    \centering
    \begin{small}\begin{sc}
    \begin{tabular}{lll|cccccccc}
    \toprule
    Model             & Alignment & Teacher           & \makecell{MT-Bench} & \makecell{MMLU}   & \makecell{ARC}   & \makecell{HellaSwag} & \makecell{Winogrande} & \makecell{GSM8K} \\ \hline
    Llama-2-13b-chat  & SFT + RLHF & \makecell[l]{Human\\annotators}                & 6.65$^\dagger$                & 54.58 & 59.81 & 82.52     & 75.93      & 34.80 \\ 
    Orca-2            & \makecell[l]{Progressive\\Training} & GPT-4 & 6.15$^\dagger$          &        60.37 & 59.73 & 79.86     & 78.22      & \textbf{48.22} \\ 
    WizardLM-13B & \makecell[l]{Evol-\\Instruct}  & GPT-4        & 7.20$^\dagger$           &        54.83  & 60.24 & 82.62     & 76.40       & 43.75  \\ \hline
    \labl       & LAB & \makecell[l]{Mixtral-8x7B-\\Instruct}                  & 7.23$^\ddagger$ &        58.89 & 61.69 & 83.15 & \textbf{79.56} & 40.11 \\ 
    \hline
    \hline
    Mistral-7B-Instruct      &  SFT  & \makecell[l]{Public\\Datasets} & 6.84$^\dagger$ & 60.37 & 63.65 & \textbf{84.76} & 76.80 & 41.85 \\ 
    Zephyr-7b-$\beta$      & SFT + DPO    & GPT-4 & 7.34$^\dagger$ &    61.07     & 63.74 & 84.19 &  78.06 & 34.04 \\ 
    \hline
    \labm      & LAB   & \makecell[l]{Mixtral-8x7B-\\Instruct} & \textbf{{7.66}}$^\ddagger$ & \textbf{64.88}  & \textbf{63.99}     & 84.37 & 78.24 & 44.58 \\ 
    \bottomrule
    \end{tabular}
    \end{sc}\end{small}
    $^\dagger$ taken from the LMSYS Chatbot Arena Leaderboard.\\
    $\ddagger$ average of 3 runs. \\
    \caption{Evaluation of LLMs with different alignment methods over a comprehensive set of benchmark metrics. Settings of each metric can be found in the main text.}\label{tab:res}
    \setlength{\tabcolsep}{6pt}
\end{table}

% \end{landscape}
Notably, in terms of MT-Bench, \textsc{\labl} performs better than the current best model fine-tuned on \textsc{Llama-2-13b} and \textsc{\labm} performs better than the current best model fine-tuned on \textsc{Mistral-7B}, achieving state-of-the-art performance in term of chatbot capability.
Importantly, out training method ensures that the model is not only good at multi-turn conversation but also maintains its knowledge or reasoning capability, as shown by the overall superior performance in the rest of the metrics.
Besides, unlike those top models that use GPT-4 as the teacher model, we achieve this performance using the open-weights \textsc{
Mixtral-8x7B-Instruct-v0.1}, which is relatively weaker teacher model at orders of magnitude less cost.

% \section{Conclusion}

\bibliography{refs}
\bibliographystyle{iclr2024_conference}

\appendix
% \section{Appendix}

\end{document}